\pdfoutput=1

\documentclass[11pt]{article}

\usepackage[]{acl}

\usepackage{times}
\usepackage{latexsym}
\usepackage{booktabs}
\usepackage{multirow}
\usepackage[table]{xcolor}
\usepackage{subfigure}
\usepackage{graphicx}
\usepackage{CJKutf8}
\usepackage{xpinyin}
\usepackage{subcaption}
\usepackage{threeparttable}
\usepackage[utf8]{inputenc}
\usepackage{makecell}
\usepackage[T1]{fontenc}

\usepackage{microtype}
\usepackage{amssymb}
\usepackage{inconsolata}

\usepackage{graphicx}
\usepackage{svg}
\usepackage{float}
\usepackage{amsmath}
\newcommand{\lz}[1]{\textcolor{blue}{{#1}}}
\title{From Word to World: Evaluate and Mitigate Culture Bias \\in LLMs via Word Association Test}

\author{
    Xunlian Dai\textsuperscript{1}, 
    Li Zhou\textsuperscript{1}\thanks{Corresponding author}, 
    Benyou Wang\textsuperscript{1}, 
    Haizhou Li\textsuperscript{1,2}
    \\
    \textsuperscript{1}The Chinese University of Hong Kong, Shenzhen \\
     \textsuperscript{2}Shenzhen Research Institute of Big Data\\
    \small{
   \href{mailto:225040015@link.cuhk.edu.cn; lizhou21@cuhk.edu.cn}{225040015@link.cuhk.edu.cn; lizhou21@cuhk.edu.cn}
 }
}

\begin{document}
\maketitle

\begin{abstract}

The human-centered word association test (WAT) serves as a cognitive proxy, revealing sociocultural variations through culturally shared semantic expectations and implicit linguistic patterns shaped by lived experiences. 
We extend this test into an LLM-adaptive, free-relation task to assess the alignment of large language models (LLMs) with cross-cultural cognition. 
To address culture preference, we propose \textit{CultureSteer}, an innovative approach that moves beyond superficial cultural prompting by embedding cultural-specific semantic associations directly within the model’s internal representation space. 
Experiments show that current LLMs exhibit significant bias toward Western (notably American) schemas at the word association level. 
In contrast, our model substantially improves cross-cultural alignment, capturing diverse semantic associations. 
Further validation on culture-sensitive downstream tasks confirms its efficacy in fostering cognitive alignment across cultures. 
This work contributes a novel methodological paradigm for enhancing cultural awareness in LLMs, advancing the development of more inclusive language technologies.\footnote{Our code and resources are publicly available at \href{https://github.com/hlt-cuhksz/CultureSteer}{https://github.com/hlt-cuhksz/CultureSteer}.}

\end{abstract}

\section{Introduction}
Word associations, 
often perceived as commonplace  neural activities~\cite{pulvermuller1999words, anderson2017predicting, khanna2024single}, are deeply rooted in  lived environments and personal experiences. The neurocognitive research~\cite{schneider2024necessity} find out that transient human associations reflect not only situational factors but also ingrained in sociocultural identities. Multilingual and cross-cultural WAT also unveil how culture modulates perceptual and interactive patterns~\cite{szalay2024subjective, garimella-etal-2017-demographic, garimella-etal-2016-identifying}.

\begin{figure}[t]

    \centering

    \includegraphics[width=1.0\linewidth]{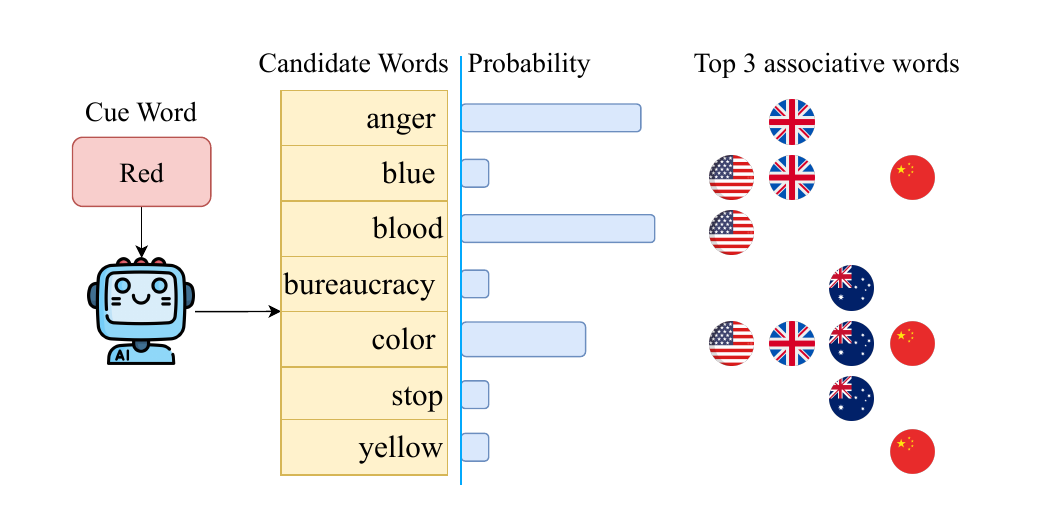}
    \caption{Cross-cultural comparison of word associations for the cue ``Red'' between LLM predictions and human responses. }
    \label{fig:introduction}
\end{figure}

For instance, as shown in Figure~\ref{fig:introduction}, when prompted with “red,” beyond general associative words such as  ``blue'' and ``color'', which reflects semantic relations, British respondents tend to activate idiomatic expressions like \textit{see red} (\textbf{anger}), while Australians are more likely to mention bureaucratic metaphors such as \textit{red tape}(\textbf{bureaucracy}). This cross-cultural divergence in word associations reveals rich cognitive-semantic information embedded within linguistic conditioning.

Multilingual LLMs exhibit persistent cultural biases, ranging from stereotypical associations~\cite{abid2021persistent, bano2025does} to value misalignments~\cite{masoud2023cultural, cao2023assessing, jiang2024can}. Prior research has largely concentrated on addressing these biases by aligning multicultural cognition through prompt-based methods~\cite{choenni2024self, wang-etal-2024-countries, sato2024reducing}, which depend on explicitly providing cultural context to elicit targeted responses. On the other hand, fine-tuning approaches~\cite{xu2024self, yao2025caredio} are knowledge-driven but still require explicit cultural settings during inference.

Building on this perspective, our approach follows the opinion that cultural competence in NLP is often best evaluated through diagnostic tasks that reveal latent cultural patterns~\cite{zhou2025culture}. By simulating human word association tasks, we directly capture latent cultural relationships and assess the semantic spaces of cultural preferences within LLMs themselves. Extending beyond existing word association datasets, our work adapts these implicit, word-level cultural patterns to the evaluation of LLMs.
Our \textbf{contributions} can be summarized as follows:

\begin{itemize}
    \item We design an LLM-adaptive, free-relation word association task and a quantitative evaluation metric to assess the cross-cultural cognitive abilities of LLMs (\S\ref{sec:task}).
    \item We introduce \textit{CultureSteer}, an innovative approach that integrates a culture-aware steering mechanism to guide semantic representations toward culturally specific spaces (\S~\ref{sec:model}).
    \item Our experiments show that LLMs exhibit cultural bias in word association tasks, while our model is able to effectively mitigate this bias, outperforming prompt-based methods (\S~\ref{sec:exp}). 
    \item Further analysis demonstrates the effectiveness of the culture-aware steering mechanism and the generalizability of the model (\S~\ref{sec:ana}).
\end{itemize}

\section{Related Work}
\paragraph{Cultural Bias and Alignment in LLMs}
Extensive research has investigated cultural biases in LLMs across domains such as value preference~\cite{benkler2023assessing, cao2023assessing, jiang2024can,durmus2023towards}, knowledge perception~\cite{palta2023fork,shen2024understanding}, and moral measurement~\cite{jinnai2024does, rao2023ethical, rao2024normad}, revealing persistent Western-centric value preferences and regionally divergent commonsense knowledge. 
To improve cultural perception, current approaches predominantly employ prompt engineering~\cite{wang-etal-2024-countries, alkhamissi-etal-2024-investigating, zhou-etal-2025-mapo, masoud-etal-2025-cultural}, explicitly embedding cultural context in prompts to evoke culturally sensitive responses. Alternatively, fine-tuning methods such as CultureBank~\cite{shi-etal-2024-culturebank}, CultureLLM~\cite{li2024culturellm}, CultureSPA~\cite{xu2024self}  and SimLLMCultureDist~\cite{cao-etal-2025-specializing} have been proposed to enhance cultural alignment in different ways.
However, these methods rely on survey data (particularly from word value surveys), data-driven SFT,  and still require culture-specific prompts for downstream tasks.


\paragraph{Word Association in LLMs}

Prior to the LLM era, previous studies have also revealed cultural preferences from the perspectives of word usage and associations~\cite{garimella-etal-2016-identifying}, and have attempted to use models, such as the Composite Skip-gram Model (C-SGM)~\cite{garimella-etal-2017-demographic}, to predict the word association process.

On the other hand, recent work adopts word association tasks to evaluate the capabilities of LLMs across various domains. For instance, uncovering gender stereotypes~\cite{abramski2024llmWA} and assessing color associative perception~\cite{fukushima2024evaluating}. These studies primarily focus on word association mappings under specific relationships, emphasizing explicit word generation and partially relying on manual interpretation during evaluation~\cite{abramski2024llm,vintar-etal-2024-human}.

In contrast, we not only extend word association datasets to four cultures, going beyond the two cultures (Chinese and English) which is focused on in the pre-LLM era, but also propose metrics that focus on the predicted probabilities of association words without manual intervention, enabling a more comprehensive evaluation of LLMs' associative capabilities.





\paragraph{Steering Control of LLMs}
Steering methods aim to control the internal activation states of LLMs at inference time to influence the generation style of their outputs~\cite{bo2025steerable}. These have been applied to reduce harmful responses~\cite{arditi2024refusal, cao2024personalized} and perform style transfer~\cite{song2025effectively, konen2024style}. Our approach introduces a steerable layer that modulates associative outputs to reflect culture-specific cognitive patterns in its lexical associations.
\section{Word Association Test Task} \label{sec:task}

Word Association Test~\cite{woodworth1911association, gough1976studying} is a psychological assessment that reveals subconscious thoughts, emotional states, or cognitive patterns through an individual's quick associative responses to stimulus words. The cultural background influences the associated words and response patterns, with different cultures often producing vastly different associations to the same stimulus, reflecting the profound impact of culture on language, thinking, and emotions.

\subsection{Human-centered task}
The WAT for human participants is conceptualized as an open-ended activity in a free-relation format, allowing for flexible and diverse associations that more accurately reflect human cognitive processes.\footnote{\url{https://smallworldofwords.org}} Before conducting the WAT, demographic information, such as age, gender, education, native language, and region, is collected for each participant. 
Given a cue word $w$, the participant $u$ is asked to provide the first word that comes to mind and may optionally add a second and third word sequentially, resulting in up to three associated words, represented as $u_{w}^{c}=\left\{ a_{1}^{u}, a_{2}^{u}, a_{3}^{u} \right\}$, where $c$ denotes the participant’s cultural background. 
Within this cultural context, the collective set of associative words for a given cue word $w$,
aggregated across all participants $U_{w}^{c}=\left\{ u_{w}^{c} \right\}$, is represented as $A_{w}^{c}=\left\{ \left( a_1,f_1 \right) , \cdots, \left( a_{n},f_n \right) \right\}$. Here, $f_i\geqslant f_{i+1}$, and $f_i$ represents the frequency with which the associative word $a_i$ appears in $U_{w}^{c}$.

\subsection{LLM-adaptive task}
Directly employing the human-centered task paradigm to conduct WAT with LLMs proves insufficient for effectively capturing variations across entire cultural groups. Inspired by~\citet{fierro-etal-2024-mulan} and \citet{zhou-etal-2025-mapo}, we adapt this task into a word prediction framework, designed as a retrieval process for candidate associative words.
Specifically, we employ a template $q$, such as ``When \{cue\_word\} is mentioned, people often associate it with the following words:'', to prompt the LLM and elicit its predictions of associated words.
The input is represented as $x = q(w)$, and the output is denoted as $o = M(q(w))$, where the generalized function $M(\cdot)$ represents the inference process of the LLM.

Unlike the human-centered task, which gathers associative words directly from participants, our method computes the probability of each candidate associative word being predicted by the LLM and ranks them based on their relative probabilities.\footnote{This can be viewed as an open-ended multi-choice question-answering task without predefined options.} 
To address potential uninformative prefixes in the LLM's next-token predictions (e.g., phrases like `I think'), we move beyond relying solely on the likelihood of a candidate associative word appearing as the immediate next token. 
Instead, we compute its probability within a fixed window of next-token predictions.
By evaluating the candidate word’s probability at each position within the prediction window, we select the maximum value as its final probability, $P_w\left( a_i \right)$.
The mathematical formulation of this process is as follows:
\begin{equation}
    P_w\left( a_i \right) =\overset{}{\max_{m\in \left[ 0,k-t \right]}}\left( \frac{1}{t}\sum_{j=0}^{t-1}{p\left( \hat{y}_{m+j}=a_{i}^{j} \right)} \right) 
\end{equation}
where $k$ is the size of the fixed prediction window, $t$ is the number of sub-tokens in the candidate word $a_i$, $a_i^j$ is the $j$-th sub-token, and $\hat{y}_{m+j}$ is the token predicted at position $m+j$. Based on these probabilities, we construct the predicted ranked list of associative words, denoted as $\hat{A}_w = [\hat{a}_0, \hat{a}_1, \dots]$, where $P_w\left( \hat{a}_i \right) \geqslant P_w\left( \hat{a}_{i+1} \right)$. 
\begin{figure*}[t]
      \centering
      \includegraphics[width=1\linewidth]{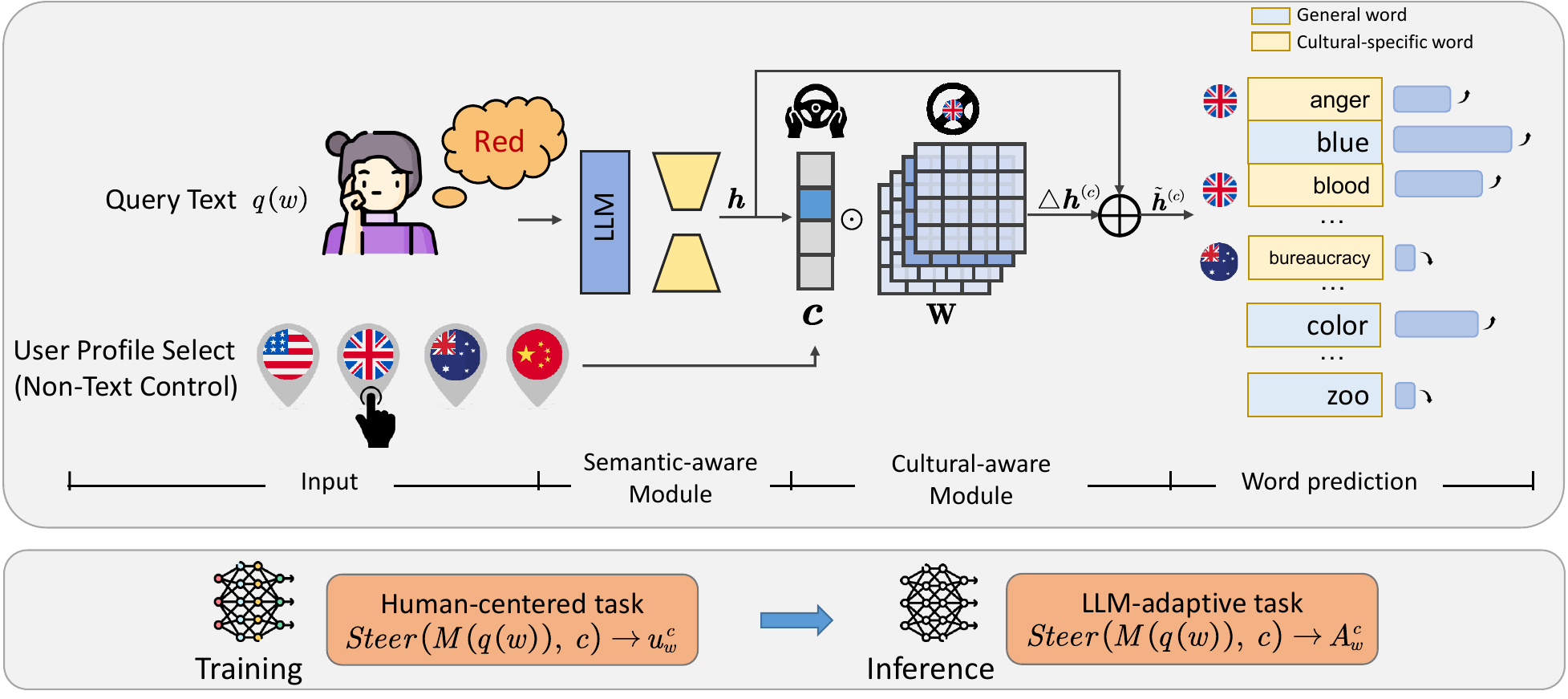}
    \caption{The framework of \textit{CultureSteer} model. Top: model pipeline;
    Bottom: training and inference process based on WAT.}
    \label{fig:framework}
\end{figure*}

\subsection{Evaluation metric} \label{sec:metric}
To evaluate the alignment between human cultural associations $A_w^c$ and the LLM's predicted associations $\hat{A}_w$, we propose a novel evaluation metric, Position-Weighted Recall (PWR@K), which is an extension of Top-K Recall (R@K).

\paragraph{Baseline Metric} 
Given that WAT in LLMs is a retrieval task involving a large number of truly relevant associative words and a ranking process close to full ranking, we adopt R@K as the baseline evaluation metric, emphasizing full coverage. Formally, it is defined as:
\begin{align}
     \begin{aligned}
     \text{R@K}&=\small{\frac{\sum_{i=1}^K{\mathbb{I} \left( \hat{a}_i\in A_{w}^{c} \right)}}{N}}\\
    &=\frac{\sum_{i=1}^N{\mathbb{I} \left( a_i\in \text{Top-K}\left( \hat{A}_{w}^{c} \right) \right)}}{N}
     \end{aligned}
\end{align}
where $N=\left| A_{w}^{c} \right|$ denotes the total number of truly relevant associative words, and $\mathbb{I}(\cdot)$ is the indicator function.

\paragraph{Proposed Metric}
While R@K effectively measures coverage, it fails to account for ranking order. In our task, higher-ranked words in $A_w^c$ signify stronger associative strength and thus hold greater importance. To address this limitation, we propose PWR@K, which incorporates positional weighting by assigning higher weights to words ranked earlier, specifically using the inverse of their position as the weight. It is formally defined as:
\begin{multline}
\text{PWR@K} = \frac{\sum_{i=1}^{N} \frac{1}{i} \cdot \mathbb{I}(a_i \in \text{Top-K}(\hat{A}_w)}{\sum_{i=1}^{N} \frac{1}{i}}
\end{multline}
where the denominator normalizes the positional weights across all $N$ positions.

A detailed comparison with other position-weighted metrics such as DCG@K is provided in Appendix~\ref{app:metric}.

\section{\textit{CultureSteer}: Modeling and Controlling Cultural Awareness in LLMs}  \label{sec:model}

Inspired by the findings of LM-Steer~\cite{han-etal-2024-word}, which suggest that, with certain assumptions, shifting styles in language models is equivalent to a linear transformation in the word embedding space, 
we extend this perspective to the cultural domain. 
Specifically, we hypothesize that different cultures define distinct semantic association spaces shaped by cultural preferences, which can be modeled through culture-specific linear transformations within the embedding space.
Based on this, we introduce the \textit{CultureSteer} model, which leverages both human-centered and LLM-adaptive WAT tasks to (1) equip LLMs with human-like word association perception, thereby enhancing their cognitive abilities, and (2) enable the word association capabilities of LLMs to align across different cultures.
The overall architecture of the model is illustrated in Figure~\ref{fig:framework}.

\paragraph{Cultural Control Paradigms}
Previous approaches primarily control cultural preferences explicitly at the input level by defining the cultural context $c$ through textual prompts. The output in these methods can be denoted as:
\begin{equation}
    o=M\left( q\left( w,c \right) \right) 
\end{equation}
In contrast, our model performs post-processing on the LLM's output, mapping the semantic representation space to a specific cultural perception space. This operation is represented as:
\begin{equation}
o = \mathrm{Steer}\left( M\left( q(w) \right), c \right)
\end{equation}
\paragraph{Cultural-Aware Steering Mechanism}
Building on this paradigm, we detail the \textit{CultureSteer} framework, focusing on how semantic representations are steered into culturally-specific spaces. 
First, we obtain the semantic representation $\boldsymbol{h}=M\left( q\left( w \right) \right)$, which encodes the general meaning without any culture-specific preferences.
To model culture-specific semantic associations, we apply a transformation to the base semantic representation $\boldsymbol{h}$. 
Specifically, the cultural preference adjustment $\bigtriangleup \boldsymbol{h}^{\left( c \right)}$ is computed as:
\begin{equation}
    \bigtriangleup \boldsymbol{h}^{\left( c \right)}=\boldsymbol{h}\cdot \left( \boldsymbol{c}\odot \mathbf{W} \right) = \boldsymbol{h}\cdot \mathbf{W}_c 
\end{equation}
Here, $\boldsymbol{c}\in \mathbb{R} ^{\left| C \right|}$ is the cultural control vector, where each element corresponds to a specific culture. In our work, we use a one-hot encoding for different cultural contexts. 
The operation $\boldsymbol{c}\odot \mathbf{W}$ selects the relevant cultural subspace from $\mathbf{W}$ called $\mathbf{W}_c$, which contains learnable parameters that encode culture-specific semantic directions.
The final culture-aware representation is then computed as:
\begin{equation}
    \tilde{\boldsymbol{h}}^{\left( c \right)}=\boldsymbol{h}+\epsilon \bigtriangleup \boldsymbol{h}^{\left( c \right)}
\end{equation}
where $\epsilon$ is a scaling factor that modulates the influence of the culture-specific adjustment.\footnote{We set $\epsilon=1e-3$ as the default steering value~\cite{han-etal-2024-word}.} The adjusted representation $ \tilde{\boldsymbol{h}}^{\left( c \right)}$ is then used to predict the associated words for a given query, considering the specified cultural context.
\paragraph{Training and Inference Process}
During training, we use the human-centered WAT pattern to train the \textit{CultureSteer} model SFT, where the true human associations are fed back into the model at each time step $t$. The loss function, which minimizes the discrepancy between the predicted associations $\tilde{a}_{t}^{u}$ and the true human responses $a_{t}^{u}$.  used is cross-entropy
During the inference phase, we employ the LLM-adaptive WAT task to evaluate the LLM's cross-cultural word association ability, using the proposed metric PWR@K.

\begin{table*}[ht]

\centering
\scalebox{0.9}{
{\small
\begin{tabular}{@{}l|lllll|llll@{}}
\toprule
                                   & \multicolumn{5}{c|}{\textbf{Llama}}                                                                                                                                      & \multicolumn{4}{c}{\textbf{Qwen}}                                                                                             \\ \midrule
                                   & PWR@K                                    & K=3                           & K=5                           & K=10                          & K=20                          & K=3                           & K=5                           & K=10                          & K=20                          \\ \midrule
                                   & \cellcolor[HTML]{D9E1F4}Baseline         & \cellcolor[HTML]{D9E1F4}12.37 & \cellcolor[HTML]{D9E1F4}19.57 & \cellcolor[HTML]{D9E1F4}29.25 & \cellcolor[HTML]{D9E1F4}42.54 & \cellcolor[HTML]{D9E1F4}15.22 & \cellcolor[HTML]{D9E1F4}19.05 & \cellcolor[HTML]{D9E1F4}26.18 & \cellcolor[HTML]{D9E1F4}35.90 \\
                                   & CultureLLM                               & 15.32                         & 19.89                         & 25.10                         & 27.76                         & 6.01                          & 7.48                          & 10.34                         & 16.15                         \\
                                   & {\color[HTML]{333333} SimLLMCultureDist} & 22.30                         & 30.94                         & 42.20                         & 50.73                         & 16.83                         & 21.89                         & 30.84                         & 42.05                         \\
                                   & CultureMerge                             & 20.29                         & 29.24                         & 40.33                         & 49.07                         & \multicolumn{1}{c}{-}         & \multicolumn{1}{c}{-}         & \multicolumn{1}{c}{-}         & \multicolumn{1}{c}{-}         \\
                                   & CultureSPA                               & 19.30                         & 27.73                         & 40.93                         & 49.71                         & \multicolumn{1}{c}{-}         & \multicolumn{1}{c}{-}         & \multicolumn{1}{c}{-}         & \multicolumn{1}{c}{-}         \\
                                   & CSP                                      & 30.30                         & 37.61                         & 48.95                         & 60.95                         & 20.34                         & 28.19                         & 39.05                         & 50.65                         \\
                                   & CCT                                      & 25.92                         & 32.98                         & 44.05                         & 55.54                         & 22.59                         & 30.51                         & 42.31                         & 54.76                         \\
\multirow{-8}{*}{\textbf{USA}}     & \textbf{CultureSteer}                    & \textbf{38.14}                & \textbf{50.30}                & \textbf{67.58}                & \textbf{80.13}                & \textbf{36.34}                & \textbf{46.06}                & \textbf{59.78}                & \textbf{72.05}                \\ \midrule
                                   & \cellcolor[HTML]{D9E1F4}Baseline         & \cellcolor[HTML]{D9E1F4}10.69 & \cellcolor[HTML]{D9E1F4}16.53 & \cellcolor[HTML]{D9E1F4}24.96 & \cellcolor[HTML]{D9E1F4}35.92 & \cellcolor[HTML]{D9E1F4}12.43 & \cellcolor[HTML]{D9E1F4}16.16 & \cellcolor[HTML]{D9E1F4}22.05 & \cellcolor[HTML]{D9E1F4}29.97 \\
                                   & CultureLLM                               & 12.68                         & 16.33                         & 20.65                         & 23.19                         & 5.68                          & 7.14                          & 9.46                          & 14.06                         \\
                                   & {\color[HTML]{333333} SimLLMCultureDist} & 19.24                         & 26.28                         & 35.00                         & 41.43                         & 14.90                         & 19.27                         & 26.04                         & 34.53                         \\
                                   & CultureMerge                             & 17.34                         & 24.95                         & 33.58                         & 41.04                         & \multicolumn{1}{c}{-}         & \multicolumn{1}{c}{-}         & \multicolumn{1}{c}{-}         & \multicolumn{1}{c}{-}         \\
                                   & CultureSPA                               & 16.46                         & 23.43                         & 33.53                         & 40.96                         & \multicolumn{1}{c}{-}         & \multicolumn{1}{c}{-}         & \multicolumn{1}{c}{-}         & \multicolumn{1}{c}{-}         \\
                                   & CSP                                      & 24.31                         & 30.64                         & 39.91                         & 49.27                         & 16.43                         & 22.14                         & 30.58                         & 39.86                         \\
                                   & CCT                                      & 21.58                         & 27.74                         & 35.70                         & 44.96                         & 18.17                         & 24.27                         & 33.54                         & 42.72                         \\
\multirow{-8}{*}{\textbf{UK}}      & \textbf{CultureSteer}                    & \textbf{29.19}                & \textbf{38.84}                & \textbf{51.45}                & \textbf{61.88}                & \textbf{28.23}                & \textbf{34.88}                & \textbf{46.10}                & \textbf{57.53}                \\ \midrule
                                   & \cellcolor[HTML]{D9E1F4}Baseline         & \cellcolor[HTML]{D9E1F4}9.53  & \cellcolor[HTML]{D9E1F4}15.62 & \cellcolor[HTML]{D9E1F4}23.25 & \cellcolor[HTML]{D9E1F4}33.67 & \cellcolor[HTML]{D9E1F4}11.87 & \cellcolor[HTML]{D9E1F4}15.07 & \cellcolor[HTML]{D9E1F4}21.35 & \cellcolor[HTML]{D9E1F4}28.41 \\
                                   & CultureLLM                               & 13.13                         & 16.82                         & 20.29                         & 22.11                         & 5.15                          & 5.97                          & 8.25                          & 13.45                         \\
                                   & {\color[HTML]{333333} SimLLMCultureDist} & 16.63                         & 22.67                         & 31.10                         & 37.80                         & 13.16                         & 17.45                         & 24.10                         & 32.88                         \\
                                   & CultureMerge                             & 15.99                         & 22.64                         & 30.64                         & 37.83                         & \multicolumn{1}{c}{-}         & \multicolumn{1}{c}{-}         & \multicolumn{1}{c}{-}         & \multicolumn{1}{c}{-}         \\
                                   & CultureSPA                               & 14.42                         & 21.20                         & 31.35                         & 37.78                         & \multicolumn{1}{c}{-}         & \multicolumn{1}{c}{-}         & \multicolumn{1}{c}{-}         & \multicolumn{1}{c}{-}         \\
                                   & CSP                                      & 23.17                         & 28.62                         & 36.78                         & 45.88                         & 15.58                         & 21.21                         & 29.02                         & 37.60                         \\
                                   & CCT                                      & 20.21                         & 25.64                         & 33.37                         & 42.41                         & 17.16                         & 22.67                         & 31.61                         & 40.63                         \\
\multirow{-8}{*}{\textbf{OC}}      & \textbf{CultureSteer}                    & \textbf{27.56}                & \textbf{35.44}                & \textbf{47.65}                & \textbf{58.77}                & \textbf{25.05}                & \textbf{32.22}                & \textbf{41.55}                & \textbf{52.22}                \\ \midrule
                                   & \cellcolor[HTML]{D9E1F4}Baseline         & \cellcolor[HTML]{D9E1F4}7.31  & \cellcolor[HTML]{D9E1F4}11.03 & \cellcolor[HTML]{D9E1F4}18.43 & \cellcolor[HTML]{D9E1F4}24.01 & \cellcolor[HTML]{D9E1F4}8.61  & \cellcolor[HTML]{D9E1F4}11.65 & \cellcolor[HTML]{D9E1F4}17.16 & \cellcolor[HTML]{D9E1F4}24.11 \\
                                   & CultureLLM                               & 0.73                          & 0.86                          & 1.01                          & 1.05                          & 4.20                          & 6.87                          & 10.56                         & 14.61                         \\
                                   & {\color[HTML]{333333} SimLLMCultureDist} & 8.42                          & 13.33                         & 19.09                         & 24.72                         & 11.78                         & 16.32                         & 23.82                         & 34.98                         \\
                                   & CultureMerge                             & 10.30                         & 15.11                         & 21.45                         & 25.77                         & \multicolumn{1}{c}{-}         & \multicolumn{1}{c}{-}         & \multicolumn{1}{c}{-}         & \multicolumn{1}{c}{-}         \\
                                   & CultureSPA                               & 10.10                         & 14.75                         & 20.57                         & 26.35                         & \multicolumn{1}{c}{-}         & \multicolumn{1}{c}{-}         & \multicolumn{1}{c}{-}         & \multicolumn{1}{c}{-}         \\
                                   & CSP                                      & 10.52                         & 15.83                         & 23.34                         & 29.81                         & 15.56                         & 21.59                         & 31.77                         & 42.81                         \\
                                   & CCT                                      & 9.71                          & 14.08                         & 21.02                         & 28.00                         & 12.32                         & 17.71                         & 26.57                         & 36.85                         \\
\multirow{-8}{*}{\textbf{CN}}      & \textbf{CultureSteer}                    & \textbf{12.77}                & \textbf{19.04}                & \textbf{27.26}                & \textbf{33.53}                & \textbf{21.78}                & \textbf{30.13}                & \textbf{44.88}                & \textbf{58.76}                \\ \midrule
                                   & \cellcolor[HTML]{D9E1F4}Baseline         & \cellcolor[HTML]{D9E1F4}9.98  & \cellcolor[HTML]{D9E1F4}15.69 & \cellcolor[HTML]{D9E1F4}23.97 & \cellcolor[HTML]{D9E1F4}34.04 & \cellcolor[HTML]{D9E1F4}12.03 & \cellcolor[HTML]{D9E1F4}15.48 & \cellcolor[HTML]{D9E1F4}21.69 & \cellcolor[HTML]{D9E1F4}29.60 \\
                                   & CultureLLM                               & 10.47                         & 13.48                         & 16.76                         & 18.53                         & 5.26                          & 6.87                          & 9.65                          & 14.57                         \\
                                   & {\color[HTML]{333333} SimLLMCultureDist} & 16.65                         & 23.31                         & 31.85                         & 38.67                         & 14.17                         & 18.73                         & 26.20                         & 36.11                         \\
                                   & CultureMerge                             & 15.98                         & 22.99                         & 31.50                         & 38.43                         & \multicolumn{1}{c}{-}         & \multicolumn{1}{c}{-}         & \multicolumn{1}{c}{-}         & \multicolumn{1}{c}{-}         \\
                                   & CultureSPA                               & 15.07                         & 21.78                         & 31.60                         & 38.70                         & \multicolumn{1}{c}{-}         & \multicolumn{1}{c}{-}         & \multicolumn{1}{c}{-}         & \multicolumn{1}{c}{-}         \\
                                   & CSP                                      & 22.08                         & 28.18                         & 37.25                         & 46.48                         & 16.98                         & 23.28                         & 32.61                         & 42.73                         \\
                                   & CCT                                      & 19.36                         & 25.11                         & 33.54                         & 42.73                         & 17.56                         & 23.79                         & 33.51                         & 43.74                         \\
\multirow{-8}{*}{\textbf{Average}} & \textbf{CultureSteer}                    & \textbf{26.92}                & \textbf{35.91}                & \textbf{48.49}                & \textbf{58.58}                & \textbf{27.85}                & \textbf{35.82}                & \textbf{48.08}                & \textbf{60.14}                \\ \bottomrule
\end{tabular}}}
\caption{Overall comparison results: (1) LLMs display cultural biases in the WAT task; (2) \textit{CultureSteer} enhances LLMs' cultural awareness and achieves the best performance.
}
\label{tab:main_result}
\end{table*}

\section{Experiment} \label{sec:exp}

\subsection{Dataset Creation}
\paragraph{Data Source}
We utilized the officially released datasets from the Small World of Words Project (SWOW)\footnote{Designed as a human-centered Word Association Task (WAT).}. For our study, we selected data in two languages: English and Mandarin Chinese, referred to as SWOW-EN~\cite{de2019small} and SWOW-ZH~\cite{li_large-scale_2024}, respectively. The SWOW-EN dataset comprises over 3 million responses collected from more than 90,000 participants across various English-speaking countries or regions, covering more than 12,000 cues. Moreover, the SWOW-ZH dataset contains over 2 million responses corresponding to 10,192 cues, with participants primarily from the Chinese mainland.

\paragraph{Culture Selection}
To ensure sufficient participant representation across different cultures, we first filter the three most prominent cultural groups from SWOW-EN: the United States (USA), the United Kingdom (UK), and Oceania (OC)\footnote{OC includes participant responses from Australia and New Zealand, which are combined due to their relatively participant numbers and shared historical background.}. Additionally, SWOW-ZH serves as the representative for China (CN). As a result, we construct datasets representing four distinct cultural groups: USA, UK, OC, and CN.

\paragraph{Data Preprocessing}
To ensure fairness in evaluating word association perception across different cultures, we utilize the Intercontinental Dictionary Series (IDS)~\cite{ids-845} to align the cue words between English and Chinese, ultimately retaining 881 cue words that are common across all four cultural groups. Given the substantial differences in participant numbers across cultures, which lead to significant variation in the number of associated words ($A_w^c$), we standardize the $A_w^c$ lists by truncating them to the size of the culture with the fewest associated words.
Using the IDS framework, we categorize these cue words into 22 semantic categories based on their conceptual meanings. Examples of the final preprocessed data and statistical details are provided in Appendix~\ref{app:paraments}, Table~\ref{tab:categories_example} and Table~\ref{tab:data_stat}. Finally, we split the cue words in each semantic category into 70\% for training and 30\% for testing.

\subsection{Experimental Setup}

\begin{figure*}[t]
\centering
    \includegraphics[width=1\textwidth]{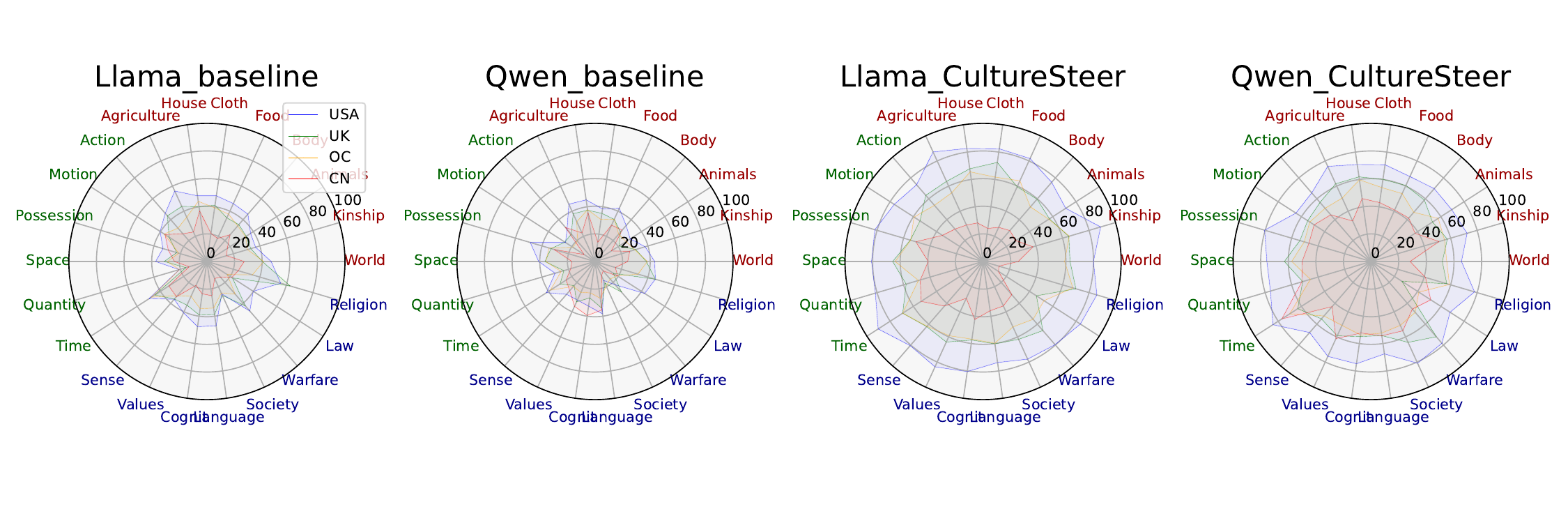}
    \caption{Fine-grained performance comparison across 22 semantic classes in the test set with PWR@20. \textcolor{red}{Red} denotes Global Knowledge, \textcolor{green}{green} denotes Perceptual Experience, and \lz{blue} denotes Cultural Ideologies. Other PWR@K (K=3, 5, 10) results are shown in Appendix~\ref{app:radars}.}
    \label{fig:main_radar}
\end{figure*}

We conduct experiments using Llama3.1-8B~\cite{grattafiori2024llama} and Qwen2.5-7B~\cite{qwen2.5} to simulate the WAT. 
First, we evaluate the base versions of these models on word association tasks using a base prompt without any cultural conditioning, aiming to assess their human-like associative cognition and the default cultural representation they exhibit. 
Subsequently, we build \textit{CultureSteer} based on these two models to enhance their cross-cultural perception capabilities. For the LLM backbone, we adopt LoRA~\cite{hu2022lora} and train the models on a single A100 GPU.

Furthermore, to comprehensively evaluate the effectiveness of our approach, we employ both prompting-based strategies and cultural-related LLMs as baseline methods for comparison.\textbf{1)} Explicit Culture-Aware Prompting Strategies: We implement two prompting approaches to examine how explicit cultural conditioning affects model performance:  First, Culture-Specific Prompt (CSP)~\cite{xu-etal-2025-self} and Cross-Culture Think (CCT). Detailed prompt templates and settings for all approaches are provided in Appendix~\ref{app:template}.
\textbf{2)} Cultural-Related LLMs: To establish comprehensive baselines, we compare against four cultural-aware language models. These include CultureSPA, Culture-merge\footnote{https://huggingface.co/surbhi21/llama3-8b-cultural-merged}, SimLLMCultureDist~\cite{cao2025specializing} and CultureLLM~\cite{li2024culturellm}\footnote{Originally trained on LLaMA2-70B, we migrate the approach to LLaMA 3.1-8B to ensure fair comparison.}.

\subsection{Overall Results}

\paragraph{LLMs Exhibit Cultural Biases in the WAT Task}
The overall comparison results are presented in Table~\ref{tab:main_result}, where the ``\colorbox[HTML]{ECF4FF}{Baseline}'' denotes the default cultural preferences exhibited by the LLMs. Both models consistently reveal a cultural preference pattern: USA > UK > OC > CN.\footnote{To demonstrate that performance differences are due to cultural factors rather than the language used in prompts, we conducted an ablation study. Details are provided in Appendix~\ref{app:translate}.} 
This suggests that, in terms of word-level cognitive abilities, LLMs exhibit a stronger inclination toward American culture, aligning with findings from prior studies~\cite{cao-etal-2023-assessing, myung2024blend, zhou-etal-2025-mapo}.
Overall, Llama demonstrates stronger human-like word-level cognitive abilities compared to Qwen, primarily due to its significantly better performance in English-speaking cultural contexts. In contrast, the performance gap between the two models is relatively small in CN.
Interestingly, when evaluating PWR@3, Qwen outperforms Llama across all cultural contexts, suggesting that Qwen excels at predicting highly associated words. However, as K increases, Llama surpasses Qwen, indicating that Llama possesses a more comprehensive associative cognition across a broader range of related words.




\paragraph{\textit{CultureSteer} Enhance LLM Cultural Awareness}
At the methodological level, \textit{CultureSteer} consistently outperforms both prompt-based methods and culture-related models across all cultural settings while some models performed worse after fine-tuning. This demonstrates that the model does not rely on explicit textual configurations but also showcases the superiority of our implicit cultural steering mechanism over traditional fine-tuning approaches that may suffer from overfitting or cultural knowledge interference. 
Interestingly, for the Llama model, the CSP method consistently outperforms the CCT method across all cultural contexts. In contrast, for the Qwen model, the CSP method performs better only in the Chinese cultural context, while the CCT method excels in the other three, particularly in English-speaking settings. This suggests that Llama is more effective with culture-specific prompts tailored to a single context, whereas Qwen demonstrates stronger cross-cultural reasoning, especially for the English-speaking environments.


\begin{table*}[ht]
\small
\centering
\resizebox{\textwidth}{!}{

\begin{tabular}{@{}lllrrrrrllllrrrrr@{}}
\toprule
TopK                                       & Model                                       & Culture & w/o $\mathbf{W}$ & $\mathbf{W}_{USA}$ & $\mathbf{W}_{UK}$ & $\mathbf{W}_{OC}$ & $\mathbf{W}_{CN}$ & \multirow{17}{*}{} & PWR@K                                      & Model                                       & Culture & w/o $\mathbf{W}$ & $\mathbf{W}_{USA}$ & $\mathbf{W}_{UK}$ & $\mathbf{W}_{OC}$ & $\mathbf{W}_{CN}$ \\ \cmidrule(r){1-8} \cmidrule(l){10-17} 
\multicolumn{1}{l|}{\multirow{8}{*}{K=3}}  & \multicolumn{1}{l|}{\multirow{4}{*}{Llama}} & USA     & 36.28            & \textbf{38.14}     & 36.21             & 36.32             & 36.32             &                    & \multicolumn{1}{l|}{\multirow{8}{*}{K=5}}  & \multicolumn{1}{l|}{\multirow{4}{*}{Llama}} & USA     & 47.96            & \textbf{50.30}     & 47.99             & 48.07             & 47.95             \\
\multicolumn{1}{l|}{}                      & \multicolumn{1}{l|}{}                       & UK      & 27.80            & 27.78              & \textbf{29.19}    & 27.96             & 27.84             &                    & \multicolumn{1}{l|}{}                      & \multicolumn{1}{l|}{}                       & UK      & 37.03            & 37.03              & \textbf{38.84}    & 37.08             & 37.03             \\
\multicolumn{1}{l|}{}                      & \multicolumn{1}{l|}{}                       & OC      & 26.48            & 26.48              & 26.36             & \textbf{27.56}    & 26.49             &                    & \multicolumn{1}{l|}{}                      & \multicolumn{1}{l|}{}                       & OC      & 33.57            & 33.55              & 33.73             & \textbf{35.44}    & 33.56             \\
\multicolumn{1}{l|}{}                      & \multicolumn{1}{l|}{}                       & CN      & 12.23            & 12.23              & 12.23             & 12.21             & \textbf{12.77}    &                    & \multicolumn{1}{l|}{}                      & \multicolumn{1}{l|}{}                       & CN      & 18.30            & 18.30              & 18.36             & 18.36             & \textbf{19.04}    \\ \cmidrule(lr){2-8} \cmidrule(l){11-17} 
\multicolumn{1}{l|}{}                      & \multicolumn{1}{l|}{\multirow{4}{*}{Qwen}}  & USA     & 37.14            & \textbf{38.04}     & 34.83             & 34.72             & 34.74             &                    & \multicolumn{1}{l|}{}                      & \multicolumn{1}{l|}{\multirow{4}{*}{Qwen}}  & USA     & 47.24            & \textbf{47.91}     & 43.87             & 43.83             & 43.89             \\
\multicolumn{1}{l|}{}                      & \multicolumn{1}{l|}{}                       & UK      & 26.45            & 26.72              & \textbf{27.84}    & 26.79             & 26.77             &                    & \multicolumn{1}{l|}{}                      & \multicolumn{1}{l|}{}                       & UK      & 34.01            & 33.21              & \textbf{35.69}    & 33.26             & 33.27             \\
\multicolumn{1}{l|}{}                      & \multicolumn{1}{l|}{}                       & OC      & \textbf{26.82}   & 25.23              & 25.20             & 25.05             & 25.19             &                    & \multicolumn{1}{l|}{}                      & \multicolumn{1}{l|}{}                       & OC      & 31.35            & 31.97              & 31.97             & \textbf{32.22}    & 32.01             \\
\multicolumn{1}{l|}{}                      & \multicolumn{1}{l|}{}                       & CN      & 21.49            & 19.54              & 19.34             & 19.35             & \textbf{21.78}    &                    & \multicolumn{1}{l|}{}                      & \multicolumn{1}{l|}{}                       & CN      & 24.81            & 24.44              & 24.15             & 24.14             & \textbf{30.13}    \\ \cmidrule(r){1-8} \cmidrule(l){10-17} 
\multicolumn{1}{l|}{\multirow{8}{*}{K=10}} & \multicolumn{1}{l|}{\multirow{4}{*}{Llama}} & USA     & 64.43            & \textbf{67.58}     & 64.50             & 64.56             & 64.40             &                    & \multicolumn{1}{l|}{\multirow{8}{*}{K=20}} & \multicolumn{1}{l|}{\multirow{4}{*}{Llama}} & USA     & 76.11            & \textbf{80.13}     & 76.20             & 76.05             & 76.08             \\
\multicolumn{1}{l|}{}                      & \multicolumn{1}{l|}{}                       & UK      & 48.88            & 49.02              & \textbf{51.45}    & 48.92             & 48.85             &                    & \multicolumn{1}{l|}{}                      & \multicolumn{1}{l|}{}                       & UK      & 58.87            & 58.97              & \textbf{61.88}    & 58.77             & 58.87             \\
\multicolumn{1}{l|}{}                      & \multicolumn{1}{l|}{}                       & OC      & 45.09            & 45.10              & 44.96             & \textbf{47.65}    & 45.05             &                    & \multicolumn{1}{l|}{}                      & \multicolumn{1}{l|}{}                       & OC      & 55.89            & 55.89              & 56.04             & \textbf{58.77}    & 55.89             \\
\multicolumn{1}{l|}{}                      & \multicolumn{1}{l|}{}                       & CN      & 26.12            & 26.12              & 26.12             & 26.12             & \textbf{27.26}    &                    & \multicolumn{1}{l|}{}                      & \multicolumn{1}{l|}{}                       & CN      & 32.09            & 32.09              & 32.09             & 32.09             & \textbf{33.53}    \\ \cmidrule(lr){2-8} \cmidrule(l){11-17} 
\multicolumn{1}{l|}{}                      & \multicolumn{1}{l|}{\multirow{4}{*}{Qwen}}  & USA     & 56.19            & \textbf{61.88}     & 56.81             & 56.83             & 56.86             &                    & \multicolumn{1}{l|}{}                      & \multicolumn{1}{l|}{\multirow{4}{*}{Qwen}}  & USA     & 68.34            & \textbf{73.94}     & 68.82             & 68.92             & 68.89             \\
\multicolumn{1}{l|}{}                      & \multicolumn{1}{l|}{}                       & UK      & 43.61            & 43.92              & \textbf{46.45}    & 44.02             & 43.87             &                    & \multicolumn{1}{l|}{}                      & \multicolumn{1}{l|}{}                       & UK      & 54.58            & 54.78              & \textbf{56.66}    & 54.74             & 54.78             \\
\multicolumn{1}{l|}{}                      & \multicolumn{1}{l|}{}                       & OC      & 41.19            & 41.28              & 41.14             & \textbf{41.55}    & 41.11             &                    & \multicolumn{1}{l|}{}                      & \multicolumn{1}{l|}{}                       & OC      & 50.73            & 50.70              & 50.73             & \textbf{52.22}    & 50.85             \\
\multicolumn{1}{l|}{}                      & \multicolumn{1}{l|}{}                       & CN      & 33.53            & 33.01              & 32.62             & 32.73             & \textbf{44.88}    &                    & \multicolumn{1}{l|}{}                      & \multicolumn{1}{l|}{}                       & CN      & 44.34            & 44.51              & 44.59             & 44.57             & \textbf{58.76}    \\ \bottomrule
\end{tabular}}
\caption{Results of controlled experiments on cultural-aware steering mechanism: performance declines without cultural preference semantic learning or with mismatched cross-cultural semantic space steering, highlighting the existence and variability of cultural semantic preferences.
}
\label{tab:cross_expert}
\end{table*}

\begin{table*}[t]
\centering
\begin{tabular}{@{}cccc|cc@{}}
\toprule
           &     & \multicolumn{2}{c|}{Llama}      & \multicolumn{2}{c}{Qwen}  \\ \midrule
           &     & Baseline       & CultureSteer   & Baseline & CultureSteer   \\
WVS & USA & \textbf{53.64} & 51.11          & 47.66    & \textbf{51.43} \\
           & UK  & 51.69          & \textbf{53.89} & 48.33    & \textbf{51.57} \\
           & OC  & 51.43          & \textbf{53.32} & 46.63    & \textbf{52.35} \\
           & CN  & 49.11          & \textbf{53.11} & 46.87    & \textbf{49.01} \\ \midrule
BLEnd      & USA & 46.29          & \textbf{50.47} & 62.73    & \textbf{68.91} \\
           & UK  & 43.18 & \textbf{48.86} & 57.89    & \textbf{62.86} \\
           & CN  & 41.65 & \textbf{47.93} & 52.84    & \textbf{56.55} \\ \bottomrule
\end{tabular}

\caption{Performance Comparison on Cross-Cultural Benchmarks: WVS and BLEnD. Bold values indicate superior performance under each model. In WVS, we evaluated the Australian Q\&A data as the result for OC, while BLEnD lacks data from Australia and New Zealand, so no evaluation was conducted.}
\label{tab:benchmark}
\end{table*}

\subsection{Fine-Grained Results} \label{sec:granular}




We visualize the fine-grained performance comparison across 22 semantic classes in the test set, as shown in Figure\ref{fig:main_radar}. 
The results reveal notable differences among baseline models in the \textit{Cognition} and \textit{Value} classes, with Qwen demonstrating value orientations more aligned with Chinese culture, while Llama exhibits stronger alignment with English cultural values.
Furthermore, based on their semantic meanings, these 22 classes are categorized into three groups: Global Knowledge, Perceptual Experience, and Cultural Ideologies.
\textit{CultureSteer} improves cultural awareness in both models with domain-dependent effectiveness. While global knowledge categories (e.g., \textit{World}, \textit{Animals}) remain difficult to align culturally, perceptual experience categories (e.g., \textit{Action}, \textit{Space}) show notable gains, especially within English-speaking cultures. Llama still exhibits gaps between English and Chinese contexts after steering, whereas Qwen achieves more balanced cross-cultural performance, nearing parity in cultural ideologies (\textit{Values}, \textit{Society}). Overall, \textit{CultureSteer} mitigates cultural biases and enhances associative abilities but cannot fully overcome inherent differences in models’ cultural perception. 


\section{Analysis} \label{sec:ana}


\subsection{Culture-Specific Semantic Learning Matters}
To validate the effectiveness of our proposed cultural-aware steering mechanism, we conduct two controlled experiments: one without cultural preference semantic learning in training and one with cross-cultural semantic space steering in inference. The results are shown in Table~\ref{tab:cross_expert}.
We observe a general decline in performance when the cultural semantic preference space is not learned or when a mismatched cultural semantic space is used for steering. This finding confirms the presence of cultural semantic preferences, highlights the differences across cultural semantic spaces, and underscores the importance of implicitly learning these preferences.
For more detail, on LLama, no significant performance differences are observed when comparing the absence of cultural preference semantic learning with the use of mismatched cross-cultural semantic space steering. However, on Qwen, mismatched cultural semantic spaces have a negative impact on performance when evaluated at $K\leq10$. Interestingly, as the evaluation range increases to $K=20$, a slight improvement in performance is observed, which suggests that broader evaluation scopes may capture additional relevant associations despite the mismatch, partially offsetting the negative effects.

\subsection{Generalization on Cultural Tasks}

To assess whether \textit{CultureSteer}'s cultural awareness and cognitive modeling capabilities benefit downstream culture-related tasks, we perform experiments on two datasets: the World Values Survey (WVS)~\cite{xu2024self}, which captures global perspectives on values and beliefs, and BLEnD~\cite{myung2024blend}, which focuses on everyday knowledge in diverse cultures.
The experimental results are shown in Table~\ref{tab:benchmark}.
\textit{CultureSteer} consistently outperforms across all cultural settings on both tasks, except for the USA culture in the WVS task, indicating that our method enhances LLMs' cultural awareness through word-level interventions. The performance gains on both WVS and BLEnD confirm that mitigating cultural bias in word association tasks establishes a foundational cognitive structure for broader cultural competence. By grounding conceptual mappings in cultural contexts through word association mechanisms, \textit{CultureSteer} activates latent cultural perception capabilities in language models, surpassing superficial mitigation of task-specific biases. This validates that word association analysis serves as an effective entry point for enhancing the generalization capacity of multicultural LLMs.



\subsection{Case Study 
}
\begin{figure}[t]
    \centering

    \begin{subfigure}
        \centering
        \includegraphics[width=0.8\linewidth]{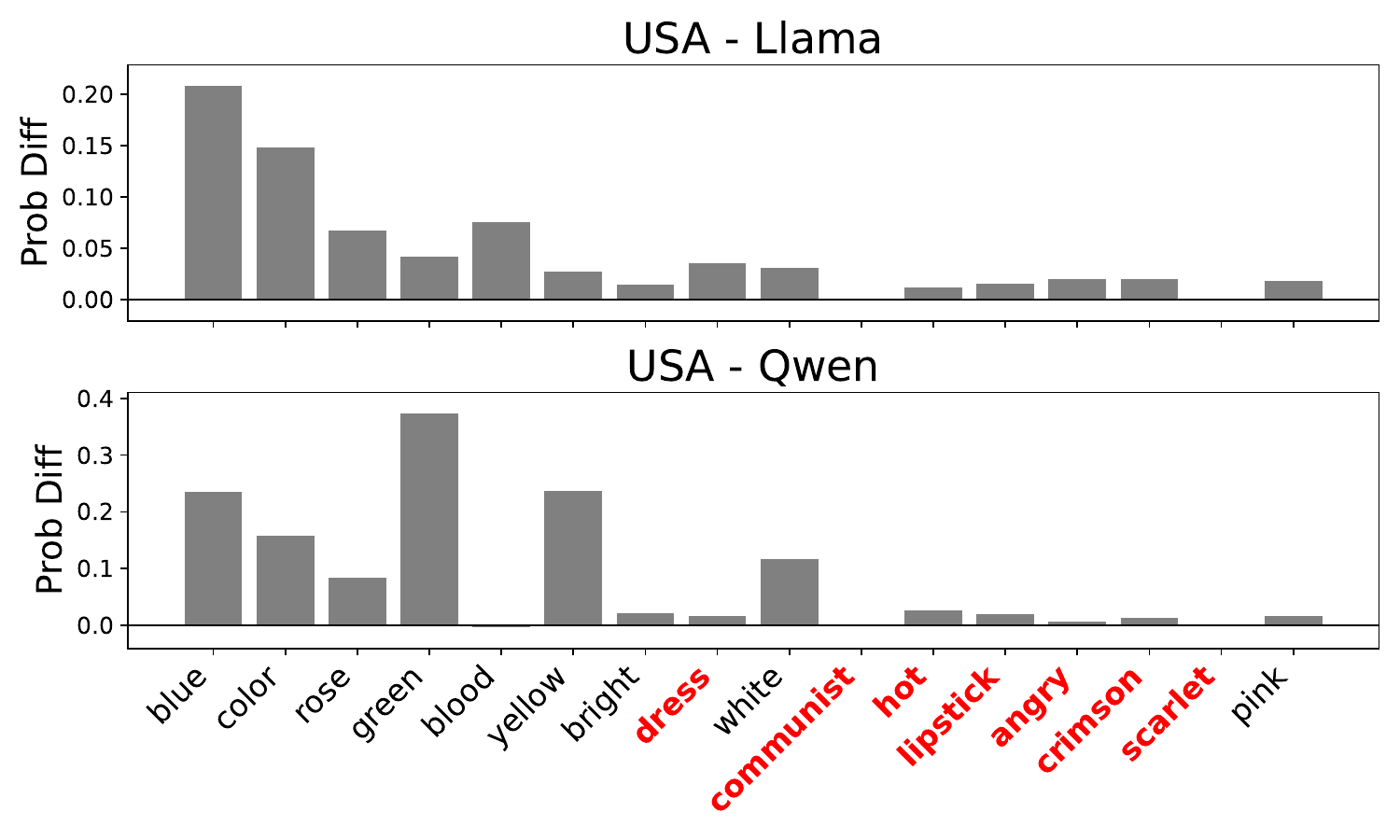}

    \end{subfigure}
    \begin{subfigure}
        \centering
        \includegraphics[width=0.8\linewidth]{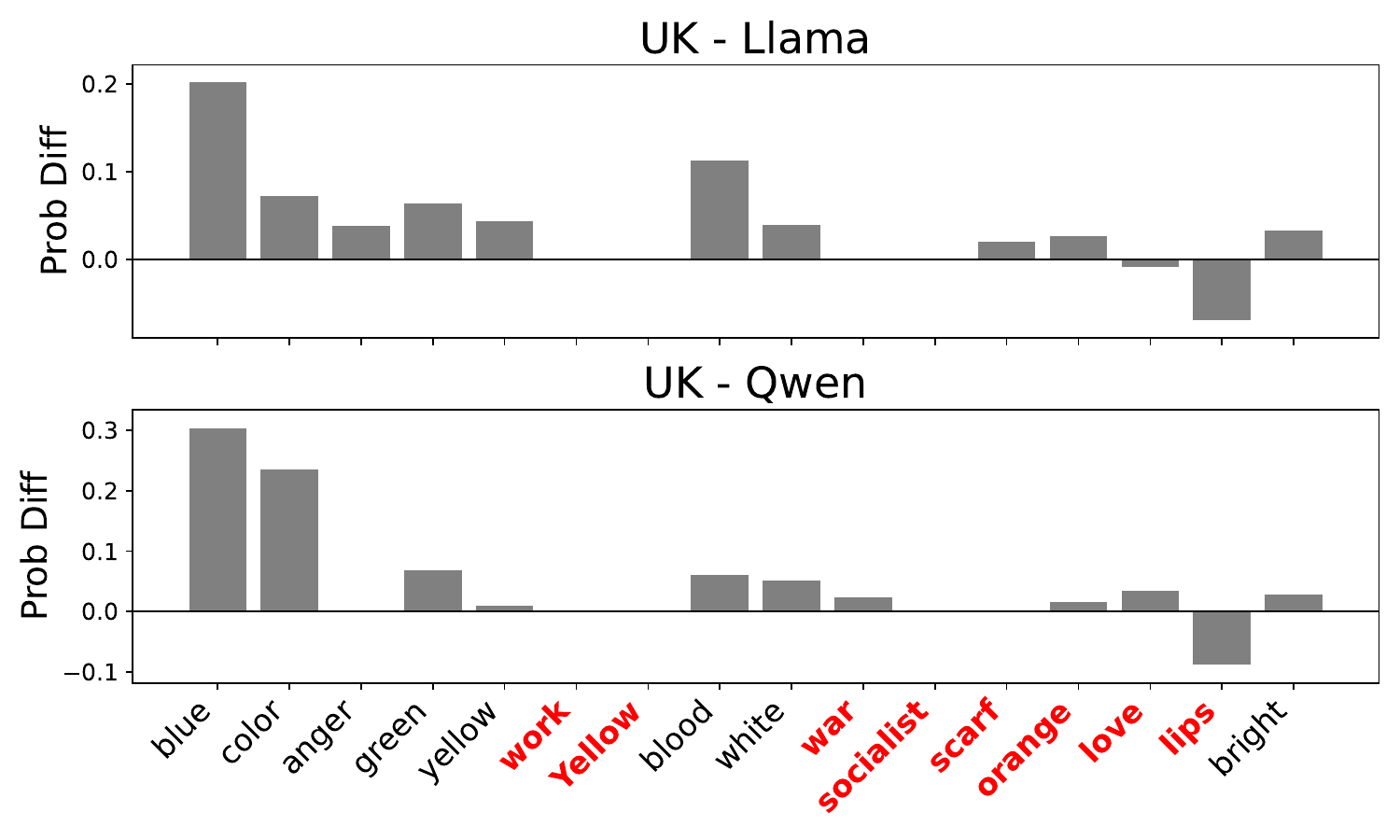}

    \end{subfigure}

    \begin{subfigure}
        \centering
        \includegraphics[width=0.8\linewidth]{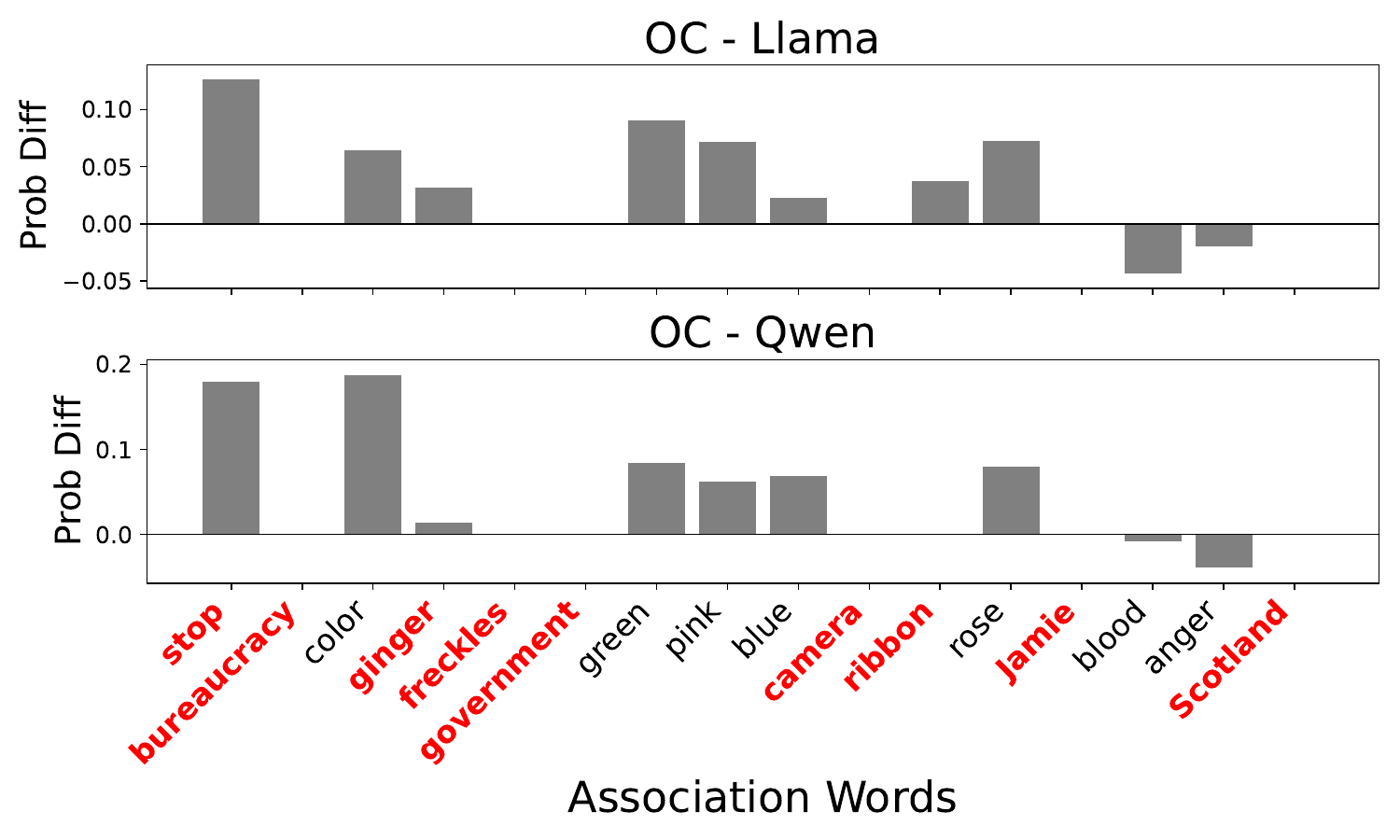}
    \end{subfigure}

    \caption{Probability differences between CultureSteer and vanilla models across USA, UK and OC. Red marks denote culture-specific words, while gray marks represent general words common across cultures.}
    \label{fig:case_result}
\end{figure}

To gain deeper insights into how \textit{CultureSteer} differentially influences the probability distributions of $\hat{A}_w^c$, we perform a targeted case study focusing on the cue word ``red'' across three English-speaking cultures: USA, UK, and Oceania (OC).\footnote{Chinese (CN) is excluded to avoid confounding linguistic differences across language families.} We compute the probability differences between \textit{CultureSteer} and the baseline LLM version for $A_{\text{red}}^c$, as illustrated in Figure~\ref{fig:case_result}. The results reveal notable probability increases for both culture-specific words (\textcolor{red}{red} marks) and general words shared across cultures. This indicates that the model learns not only universal word associations but also culture-specific ones, validating the effectiveness of the \textit{CultureSteer}.

\section{Conclusion}

Our research aims to utilize the insights from WAT to evaluate and mitigate cultural biases in LLMs. 
We proposed a novel metrics for assessing the association abilities of LLMs through word prediction task. Additionally, we introduced \textit{CultureSteer} that leverages word association patterns to align LLMs' cultural cognition by guiding sematic representations toward culturally specific spaces and validated effectiveness of this method in downstream tasks.
Although we did not explicitly prescribe the specific associative logic underlying word associations, we instead enabled the model to infer these connections through contextual prompts. Remarkably, our training methodology successfully elicited the model's cross-cultural word association capabilities. 
Future research could concentrate on refining these associative patterns to further augment models' cultural sensitivity and semantic comprehension.

\section*{Limitation}
Our research is constrained by its focus on a limited number of languages and cultures when examining word association differences, primarily due to the reliance on word association test datasets sourced from human participants, which poses significant challenges to scalability. Moreover, our control vector mechanism treats cultures as independent entities, without adequately considering potential relationships and interactions between them. Future work could address these limitations by developing scalable methods for collecting word association data, expanding the linguistic and cultural scope, and incorporating inter-cultural connections to provide a more comprehensive and nuanced perspective.

\section*{Acknowledgments}
This work is supported by Shenzhen Science and Technology Program (Shenzhen Key Laboratory, Grant No. ZDSYS20230626091302006), Program for Guangdong Introducing Innovative and Entrepreneurial Teams (Grant No. 2023ZT10X044), Shenzhen Stability Science Program 2023, Shenzhen Key Lab of Multi-Modal Cognitive Computing, Shenzhen Science and Technology Research Fund (Fundamental Research Key Project, Grant No.JCYJ20220818103001002), the International Science and Technology Cooperation Center, Ministry of Science and Technology of China (under grant 2024YFE0203000), the Internal Project of Shenzhen Research Institute of Big Data.

\bibliography{custom,anthology}

\appendix

\section{Data Preparation} \label{app:data}
\subsection{Few example of IDS chapters} \label{app:example}

We list the whole 22 chapters of IDS and a few example words, shown in Table \ref{tab:categories_example}

\begin{table}[ht]
\centering
\scalebox{0.8}{
\begin{tabular}{cc} 
\toprule
\textbf{Chapter}            & \textbf{Examples(English)}       \\
\midrule
Physical world              & flame, ice, light     \\
Kinship                     & they, girl, male \\
Animals                     & bee, ant, fox      \\
The body                    & blind, wound, dream   \\
Food and drink              & fruit, mill, dish      \\
Clothing and grooming       & headband, coat, tailor \\
The house                   & brick, window, yard                       \\
Agricultural and vegetation & crop, plow, garden                    \\
Action and technology       & wash, untie, glass                            \\
Motion                      & harbor, carriage, road                      \\
Possession                  & market, trade, tax                          \\
Spatial relations           & wide, south, far                         \\
Quantity                    & second, some, more                          \\
Time                        & often, age, summer                           \\
Sense perception            & see, color, heavy                          \\
Emotion and values          & joyful, love, hate                           \\
Cognition                   & know, learn, pupil                      \\
Speech and Language         & ask, refuse, paper                        \\
Society and politics        & master, friend, custom                            \\
Warfare and hunting         & fight, soldier, shoot                          \\
Law                         & court, accuse, steal                      \\
Religion and belief         & bless, heaven, ghost                       \\
\bottomrule
\end{tabular}}

\caption{The 22 chapters names with few sample words}
\label{tab:categories_example}
\end{table}

\subsection{Data Cleaning} \label{app:data_clean}
We filter SWOW with IDS by the following steps:

\begin{itemize}
    \item For one-to-many alignments in IDS, we retain only the first word in the list of multiple terms.
    \item Align the cue words between English and Chinese in IDS cross-culture word paris
    \item Remove abnormal characters and meaningless responses, such as \texttt{'\#Missing'} or \texttt{'?'}.
    \item Filter out responses containing multiple words in English and those with more than four characters in Chinese.
    \item Truncate the association words for each cue word to the number of association words corresponding to the culture with fewest association words.
\end{itemize}

\subsection{Data Overview} \label{app:data_stat}
The data overview(frequency) of our dataset has been shown in Table~\ref{tab:data_stat}. By truncating the association word, we ensure that the number of tuples for each association sorting $A_w$ corresponding to each cue word is consistent across cultures.

\begin{table}[ht]
\centering
\scalebox{0.76}{
\begin{tabular}{@{}lrrrrrr@{}}
\toprule
\multirow{2}{*}{\textbf{Chap}} & \multirow{2}{*}{\textbf{Cue}} & \multirow{2}{*}{\textbf{\begin{tabular}[c]{@{}r@{}}Asso\\ (avg.)\end{tabular}}} & \multicolumn{4}{c}{\textbf{Users}}                              \\ \cmidrule(l){4-7} 
                               &                               &                                                                                 & \textbf{USA}   & \textbf{UK}   & \textbf{OC}   & \textbf{CN}    \\ \midrule
\textbf{World}                 & 52                            & 15.96                                                                           & 3895           & 645           & 513           & 1995           \\
\textbf{Kinship}               & 34                            & 14.21                                                                           & 2246           & 342           & 266           & 1282           \\
\textbf{Animals}               & 65                            & 14.2                                                                            & 3416           & 706           & 459           & 2420           \\
\textbf{Body}                  & 92                            & 13.12                                                                           & 4988           & 897           & 679           & 3462           \\
\textbf{Food}                  & 53                            & 14.25                                                                           & 3046           & 519           & 379           & 2066           \\
\textbf{Cloth}                 & 44                            & 12.95                                                                           & 2296           & 450           & 273           & 1631           \\
\textbf{House}                 & 31                            & 13.81                                                                           & 1784           & 267           & 234           & 1205           \\
\textbf{Agriculture}           & 36                            & 13.81                                                                           & 2285           & 409           & 270           & 1337           \\
\textbf{Action}                & 39                            & 14.08                                                                           & 2336           & 362           & 353           & 1436           \\
\textbf{Motion}                & 47                            & 15.36                                                                           & 2910           & 465           & 488           & 1617           \\
\textbf{Possession}            & 33                            & 12.97                                                                           & 1841           & 237           & 313           & 1175           \\
\textbf{Space}                 & 59                            & 12.8                                                                            & 3544           & 406           & 529           & 1970           \\
\textbf{Quantity}              & 29                            & 12.69                                                                           & 1456           & 166           & 296           & 924            \\
\textbf{Time}                  & 42                            & 12.05                                                                           & 2711           & 289           & 370           & 1455           \\
\textbf{Sense}                 & 34                            & 15.38                                                                           & 1975           & 292           & 426           & 1337           \\
\textbf{Values}                & 42                            & 13.17                                                                           & 2243           & 346           & 319           & 1670           \\
\textbf{Cognit}                & 38                            & 13.45                                                                           & 2299           & 277           & 323           & 1332           \\
\textbf{Language}              & 31                            & 14.13                                                                           & 1978           & 277           & 271           & 1073           \\
\textbf{Society}               & 27                            & 14.19                                                                           & 1353           & 229           & 183           & 957            \\
\textbf{Warfare}               & 26                            & 14.46                                                                           & 1428           & 277           & 158           & 940            \\
\textbf{Law}                   & 16                            & 15.19                                                                           & 1078           & 135           & 135           & 682            \\
\textbf{Religion}              & 11                            & 14.36                                                                           & 592            & 104           & 72            & 440            \\ \midrule
\textbf{Overall}               & \textbf{881}                  & \textbf{13.94}                                                                  & \textbf{51700} & \textbf{8097} & \textbf{7309} & \textbf{32406} \\ \bottomrule
\end{tabular}}
\caption{The data overview(frequency) of our dataset.}
\label{tab:data_stat}
\end{table}
\section{Metric Discussing} \label{app:metric}
We provide a detailed comparison with Discounted Cumulative Gain at K (DCG@K), a widely-used position-weighted evaluation metric in information retrieval.

\paragraph{DCG@K Formulation}
DCG@K applies a logarithmic discount to the relevance scores based on their ranking positions, with the intuition that items ranked lower should contribute less to the overall score. 
For our Word Association Task (WAT), we adapt DCG@K as follows. Given the ground truth associative words $A_w^c = \{a_1, a_2, \ldots, a_N\}$ ranked by their associative strength, and the LLM's predicted ranking $\hat{A}_w$, DCG@K is defined as:

\begin{align}
\text{DCG@K} = \sum_{i=1}^{K} \frac{\text{rel}(i)}{\log_2(i+1)}
\end{align}

where $\text{rel}(i)$ represents the relevance score of the item at position $i$ in the predicted ranking. In our adaptation, we define:

\begin{align}
\text{rel}(i) = \begin{cases}
\frac{N - \text{rank}(\hat{a}_i) + 1}{N} & \text{if } \hat{a}_i \in A_w^c \\
0 & \text{otherwise}
\end{cases}
\end{align}

where $\text{rank}(\hat{a}_i)$ denotes the ground truth ranking position of the predicted word $\hat{a}_i$ in $A_w^c$. This formulation assigns higher relevance scores to words that appear earlier in the ground truth ranking.

\paragraph{Experimental Results with DCG@K}
We re-evaluate all our main experiments using DCG@K and present the results in Table~\ref{tab:dcg_result}. The experimental setup and datasets remain identical to those reported in the main paper.

The results obtained using NDCG@K are consistent with those derived from our proposed PWR@K metric. Both metrics demonstrate similar trends across different models, cultural contexts, and experimental conditions. This consistency validates that our key findings and conclusions are robust to the choice of position-weighted evaluation metric.

\begin{table*}[ht]
\centering
\small
\begin{tabular}{@{}l|lllll|llll@{}}
\toprule
                                   & \multicolumn{5}{c|}{\textbf{Llama}}                                                                                                                                  & \multicolumn{4}{c}{\textbf{Qwen}}                                                                                         \\ \midrule
                                   & PWR@K                                    & K=3                          & K=5                          & K=10                         & K=20                         & K=3                          & K=5                          & K=10                         & K=20                         \\ \midrule
                                   & \cellcolor[HTML]{D9E1F4}Baseline         & \cellcolor[HTML]{D9E1F4}0.88 & \cellcolor[HTML]{D9E1F4}1.17 & \cellcolor[HTML]{D9E1F4}1.54 & \cellcolor[HTML]{D9E1F4}1.78 & \cellcolor[HTML]{D9E1F4}0.82 & \cellcolor[HTML]{D9E1F4}1.00 & \cellcolor[HTML]{D9E1F4}1.23 & \cellcolor[HTML]{D9E1F4}1.51 \\
                                   & CultureLLM                               & 0.78                         & 0.93                         & 1.08                         & 1.12                         & 0.29                         & 0.35                         & 0.44                         & 0.54                         \\
                                   & {\color[HTML]{333333} SimLLMCultureDist} & 0.98                         & 1.27                         & 1.63                         & 1.87                         & 0.82                         & 1.00                         & 1.23                         & 1.51                         \\
                                   & CultureMerge                             & 0.92                         & 1.20                         & 1.57                         & 1.79                         & \multicolumn{1}{c}{-}        & \multicolumn{1}{c}{-}        & \multicolumn{1}{c}{-}        & \multicolumn{1}{c}{-}        \\
                                   & CultureSPA                               & 0.87                         & 1.15                         & 1.53                         & 1.76                         & \multicolumn{1}{c}{-}        & \multicolumn{1}{c}{-}        & \multicolumn{1}{c}{-}        & \multicolumn{1}{c}{-}        \\
                                   & CSP                                      & 1.04                         & 1.31                         & 1.65                         & 1.90                         & 1.02                         & 1.22                         & 1.50                         & 1.77                         \\
                                   & CCT                                      & 0.83                         & 1.06                         & 1.33                         & 1.49                         & 1.04                         & 1.21                         & 1.46                         & 1.72                         \\
\multirow{-8}{*}{\textbf{USA}}     & \textbf{CultureSteer}                    & \textbf{1.49}                & \textbf{1.87}                & \textbf{2.36}                & \textbf{2.66}                & \textbf{1.38}                & \textbf{1.71}                & \textbf{2.13}                & \textbf{2.40}                \\ \midrule
                                   & \cellcolor[HTML]{D9E1F4}Baseline         & \cellcolor[HTML]{D9E1F4}0.94 & \cellcolor[HTML]{D9E1F4}1.18 & \cellcolor[HTML]{D9E1F4}1.35 & \cellcolor[HTML]{D9E1F4}0.68 & \cellcolor[HTML]{D9E1F4}0.70 & \cellcolor[HTML]{D9E1F4}0.82 & \cellcolor[HTML]{D9E1F4}1.00 & \cellcolor[HTML]{D9E1F4}1.20 \\
                                   & CultureLLM                               & 0.70                         & 0.81                         & 0.85                         & 0.62                         & 0.25                         & 0.30                         & 0.37                         & 0.45                         \\
                                   & {\color[HTML]{333333} SimLLMCultureDist} & 1.03                         & 1.28                         & 1.43                         & 0.77                         & 0.69                         & 0.82                         & 1.00                         & 1.20                         \\
                                   & CultureMerge                             & 0.97                         & 1.22                         & 1.38                         & 0.74                         & \multicolumn{1}{c}{-}        & \multicolumn{1}{c}{-}        & \multicolumn{1}{c}{-}        & \multicolumn{1}{c}{-}        \\
                                   & CultureSPA                               & 0.95                         & 1.20                         & 1.37                         & 0.68                         & \multicolumn{1}{c}{-}        & \multicolumn{1}{c}{-}        & \multicolumn{1}{c}{-}        & \multicolumn{1}{c}{-}        \\
                                   & CSP                                      & 1.00                         & 1.24                         & 1.41                         & 0.81                         & 0.79                         & 0.95                         & 1.16                         & 1.35                         \\
                                   & CCT                                      & 0.85                         & 1.04                         & 1.17                         & 0.68                         & 0.81                         & 0.95                         & 1.15                         & 1.33                         \\
\multirow{-8}{*}{\textbf{UK}}      & \textbf{CultureSteer}                    & \textbf{1.38}                & \textbf{1.71}                & \textbf{1.94}                & \textbf{1.16}                & \textbf{1.03}                & \textbf{1.27}                & \textbf{1.55}                & \textbf{1.77}                \\ \midrule
                                   & \cellcolor[HTML]{D9E1F4}Baseline         & \cellcolor[HTML]{D9E1F4}0.68 & \cellcolor[HTML]{D9E1F4}0.91 & \cellcolor[HTML]{D9E1F4}1.18 & \cellcolor[HTML]{D9E1F4}1.33 & \cellcolor[HTML]{D9E1F4}0.70 & \cellcolor[HTML]{D9E1F4}0.83 & \cellcolor[HTML]{D9E1F4}1.01 & \cellcolor[HTML]{D9E1F4}1.20 \\
                                   & CultureLLM                               & 0.62                         & 0.72                         & 0.83                         & 0.86                         & 0.28                         & 0.33                         & 0.40                         & 0.49                         \\
                                   & {\color[HTML]{333333} SimLLMCultureDist} & 0.77                         & 0.98                         & 1.22                         & 1.39                         & 0.70                         & 0.83                         & 1.01                         & 1.20                         \\
                                   & CultureMerge                             & 0.74                         & 0.95                         & 1.20                         & 1.37                         & \multicolumn{1}{c}{-}        & \multicolumn{1}{c}{-}        & \multicolumn{1}{c}{-}        & \multicolumn{1}{c}{-}        \\
                                   & CultureSPA                               & 0.68                         & 0.90                         & 1.17                         & 1.32                         & \multicolumn{1}{c}{-}        & \multicolumn{1}{c}{-}        & \multicolumn{1}{c}{-}        & \multicolumn{1}{c}{-}        \\
                                   & CSP                                      & 0.81                         & 1.00                         & 1.22                         & 1.39                         & 0.82                         & 0.98                         & 1.19                         & 1.37                         \\
                                   & CCT                                      & 0.68                         & 0.85                         & 1.04                         & 1.16                         & 0.79                         & 0.93                         & 1.10                         & 1.28                         \\
\multirow{-8}{*}{\textbf{OC}}      & \textbf{CultureSteer}                    & \textbf{1.16}                & \textbf{1.44}                & \textbf{1.77}                & \textbf{1.99}                & \textbf{1.04}                & \textbf{1.28}                & \textbf{1.56}                & \textbf{1.79}                \\ \midrule
                                   & \cellcolor[HTML]{D9E1F4}Baseline         & \cellcolor[HTML]{D9E1F4}0.62 & \cellcolor[HTML]{D9E1F4}0.80 & \cellcolor[HTML]{D9E1F4}1.01 & \cellcolor[HTML]{D9E1F4}1.16 & \cellcolor[HTML]{D9E1F4}0.77 & \cellcolor[HTML]{D9E1F4}0.98 & \cellcolor[HTML]{D9E1F4}1.26 & \cellcolor[HTML]{D9E1F4}1.58 \\
                                   & CultureLLM                               & 0.06                         & 0.06                         & 0.07                         & 0.07                         & 0.29                         & 0.38                         & 0.52                         & 0.65                         \\
                                   & {\color[HTML]{333333} SimLLMCultureDist} & 0.58                         & 0.76                         & 0.96                         & 1.11                         & 0.77                         & 0.98                         & 1.26                         & 1.58                         \\
                                   & CultureMerge                             & 0.59                         & 0.76                         & 0.95                         & 1.08                         & \multicolumn{1}{c}{-}        & \multicolumn{1}{c}{-}        & \multicolumn{1}{c}{-}        & \multicolumn{1}{c}{-}        \\
                                   & CultureSPA                               & 0.62                         & 0.79                         & 1.00                         & 1.14                         & \multicolumn{1}{c}{-}        & \multicolumn{1}{c}{-}        & \multicolumn{1}{c}{-}        & \multicolumn{1}{c}{-}        \\
                                   & CSP                                      & 0.80                         & 1.03                         & 1.30                         & 1.47                         & 1.13                         & 1.45                         & 1.94                         & 2.35                         \\
                                   & CCT                                      & 0.81                         & 1.01                         & 1.24                         & 1.39                         & 0.94                         & 1.24                         & 1.68                         & 2.06                         \\
\multirow{-8}{*}{\textbf{CN}}      & \textbf{CultureSteer}                    & \textbf{1.06}                & \textbf{1.35}                & \textbf{1.69}                & \textbf{1.87}                & \textbf{1.45}                & \textbf{1.88}                & \textbf{2.54}                & \textbf{3.06}                \\ \midrule
                                   & \cellcolor[HTML]{D9E1F4}Baseline         & \cellcolor[HTML]{D9E1F4}0.90 & \cellcolor[HTML]{D9E1F4}0.91 & \cellcolor[HTML]{D9E1F4}1.18 & \cellcolor[HTML]{D9E1F4}1.37 & \cellcolor[HTML]{D9E1F4}0.86 & \cellcolor[HTML]{D9E1F4}0.86 & \cellcolor[HTML]{D9E1F4}1.06 & \cellcolor[HTML]{D9E1F4}1.29 \\
                                   & CultureLLM                               & 0.71                         & 0.60                         & 0.70                         & 0.73                         & 0.33                         & 0.32                         & 0.40                         & 0.50                         \\
                                   & {\color[HTML]{333333} SimLLMCultureDist} & 0.99                         & 0.97                         & 1.22                         & 1.41                         & 0.85                         & 0.86                         & 1.06                         & 1.29                         \\
                                   & CultureMerge                             & 0.95                         & 0.93                         & 1.19                         & 1.37                         & \multicolumn{1}{c}{-}        & \multicolumn{1}{c}{-}        & \multicolumn{1}{c}{-}        & \multicolumn{1}{c}{-}        \\
                                   & CultureSPA                               & 0.90                         & 0.91                         & 1.17                         & 1.36                         & \multicolumn{1}{c}{-}        & \multicolumn{1}{c}{-}        & \multicolumn{1}{c}{-}        & \multicolumn{1}{c}{-}        \\
                                   & CSP                                      & 1.02                         & 1.03                         & 1.29                         & 1.50                         & 1.00                         & 1.07                         & 1.33                         & 1.61                         \\
                                   & CCT                                      & 0.84                         & 0.89                         & 1.11                         & 1.27                         & 0.98                         & 1.01                         & 1.24                         & 1.50                         \\
\multirow{-8}{*}{\textbf{Average}} & \textbf{CultureSteer}                    & \textbf{1.45}                & \textbf{1.44}                & \textbf{1.80}                & \textbf{2.07}                & \textbf{1.31}                & \textbf{1.43}                & \textbf{1.78}                & \textbf{2.13}                \\ \bottomrule
\end{tabular}

\caption{
Re-evaluated experimental results using DCG@K show consistent performance trends with our proposed PWR@K metric across models and conditions, confirming the validity and robustness of PWR@K for assessing ranking effectiveness.
}
\label{tab:dcg_result}

\end{table*}
\section{Experiment Settings}
\subsection{Template Using} \label{app:template}
The templates used in this work are shown in Table~\ref{tab:templates}. Notably, the CN-t condition also employs English prompts.

\begin{table*}[h!]
\small
\resizebox{\textwidth}{!}{
\renewcommand{\arraystretch}{2} 
\begin{tabular}{ccc} \toprule
\multicolumn{1}{l}{Language} & Mode & Template                                                                                         \\ \midrule
\multirow{4}{*}{EN,CN-t}          & Base & When "\{cue\_word\}" is mentioned, people often think of the following words:                    \\ \cmidrule{2-3}
                             & CSP  & \makecell{\textcolor{red}{You are a person with \{culture\} cultural background.} \textcolor{orange}{You will be performing a word association task.} \\ \textcolor{orange}{Please directly answer the association word.} \\ When "\{cue\_word\}" is mentioned, people often think of the following words:}                                          \\ \cmidrule{2-3}
                             & CCT  & \makecell{\textcolor{red}{You are a person with \{culture\} cultural background.} \textcolor{orange}{You will be performing a word association task.} \\   \textcolor{orange}{Please directly answer the association word.}  \\ \textcolor{blue}{Before you respond, think about how \{culture\} culture is different from \{cultures\} cultures.} \\ When "\{cue\_word\}" is mentioned, people often think of the following words} \\ \midrule
\multirow{4}{*}{CN}          & Base & \begin{CJK*}{UTF8}{gbsn}当提起"\{cue\_word\}",人们往往会想到的词是: \end{CJK*}                                                                  \\ \cmidrule{2-3}
                             & CSP  & \makecell{\begin{CJK*}{UTF8}{gbsn}\textcolor{red}{你是一个中国文化背景的人。}\end{CJK*} \\  \begin{CJK*}{UTF8}{gbsn}\textcolor{orange}{你将进行词联想任务，请直接说出你联想到的词。} \end{CJK*} \\ \begin{CJK*}{UTF8}{gbsn}当提起"\{cue\_word\}",人们往往会想到的词是: \end{CJK*} }                                                                               \\ \cmidrule{2-3}
                             & CCT  & \makecell{ \begin{CJK*}{UTF8}{gbsn}\textcolor{red}{你是一个中国文化背景的人。}\end{CJK*}   \begin{CJK*}{UTF8}{gbsn}\textcolor{orange}{你将进行词联想任务， 请直接说出你联想到的词。} \end{CJK*} \\  \begin{CJK*}{UTF8}{gbsn}\textcolor{blue}{在回答之前, 请你注意中国文化与美国、英国、大洋洲文化的不同。}\end{CJK*} \\ \begin{CJK*}{UTF8}{gbsn}当提起"\{cue\_word\}",人们往往会想到的词是: \end{CJK*}  }                                                               \\ \bottomrule
\end{tabular}
}

\caption{Templates for different tasks and regions, where \textcolor{red}{red} denotes additional cultural information, \textcolor{blue}{blue} indicates cross-cultural thinking instructions, and \textcolor{orange}{orange} marks task descriptions aimed at preventing LLMs from generating redundant responses.}
\label{tab:templates}
\end{table*}

\subsection{Hyperparaments} \label{app:paraments}
The detailed setting of hyperparaments are shown in Table~\ref{tab:parament}.
\begin{table}[h!]
\centering
\begin{tabular}{@{}lll@{}}
\toprule
Phase   & Hyperparameters & Value \\ \midrule
Train   & learn rate      & 1e-4  \\
        & epsilon         & 1e-3  \\
        & max length      & 64    \\
        & l2 rate         & 0     \\
        & LoRA rank       & 8     \\
        & LoRA alpha      & 16    \\
        & LoRA dropout    & 0.05  \\
        & batch size      & 8     \\
        & epoch           & 5     \\ \midrule
Generate & max new tokens  & 5     \\
        & temperature     & 1     \\
        & epsilon         & 1e-3  \\ \bottomrule
\end{tabular}
\caption{Relative hyperparameters}
\label{tab:parament}
\end{table}

\section{Language Ablation Study} \label{app:translate}
\begin{table}[htb]
\centering
\small
\begin{tabular}{@{}cllll@{}}
\toprule
\multicolumn{5}{c}{Llama(baseline)}                                                                                         \\ \midrule
\multicolumn{1}{l}{PWR@K} & \multicolumn{1}{c}{3} & \multicolumn{1}{c}{5} & \multicolumn{1}{c}{10} & \multicolumn{1}{c}{20} \\ \midrule
\textbf{CN}               & \textbf{7.31}         & \textbf{11.03}        & \textbf{18.43}         & \textbf{24.01}         \\
\textbf{CN-t}             & 5.51                  & 8.79                  & 15.62                  & 23.75                  \\ \midrule
\multicolumn{5}{c}{Qwen(baseline)}                                                                                          \\ \midrule
\textbf{CN}               & \textbf{8.61}         & \textbf{11.65}        & \textbf{17.16}         & \textbf{24.11}         \\
\textbf{CN-t}             & 5.55                  & 7.21                  & 10.33                  & 15.26                  \\ \bottomrule
\end{tabular}
\caption{Comparison results after translating Chinese prompts and lexical items into English. Bold values indicate stronger activation of word association capabilities in Chinese linguistic contexts.}
\label{tab:translate}
\end{table}


In \S~\ref{sec:exp}, we performed an ablation study introducing a CN-t (Chinese-to-English translated) group to investigate the impact of linguistic differences on our results, controlling for language while preserving English context. Chinese-associated words were translated using MUSE bilingual dictionaries~\cite{lample2017unsupervised}, with Opus-MT-zh-en~\cite{tiedemann2024democratizing} covering uncovered pairs. Results in Table~\ref{tab:translate} suggest that Chinese contexts more effectively activate Chinese word association capabilities, indicating that the main experimental performance differences stem from cultural rather than linguistic factors.

\section{Granular Performance}
\label{app:radars}

\begin{figure*}[ht]
    \centering
    \subfigure[Performance in PWR@3]{
        \includegraphics[width=1\textwidth]{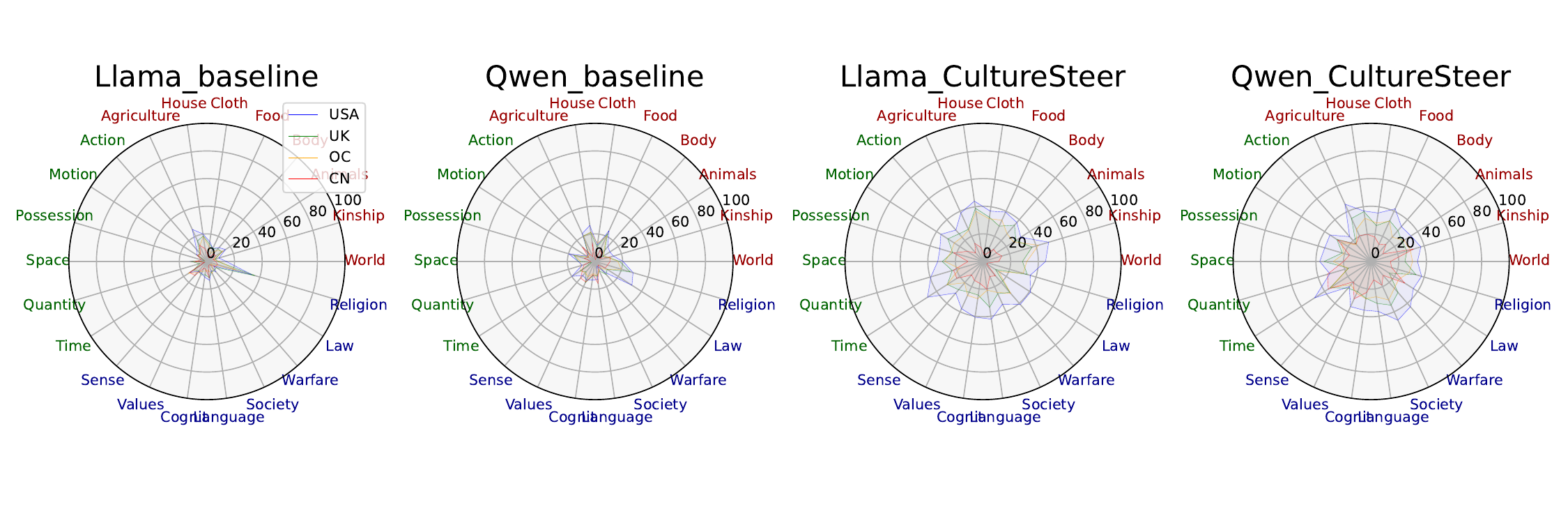}
    }
    \subfigure[Performance in PWR@5]{
        \includegraphics[width=1\textwidth]{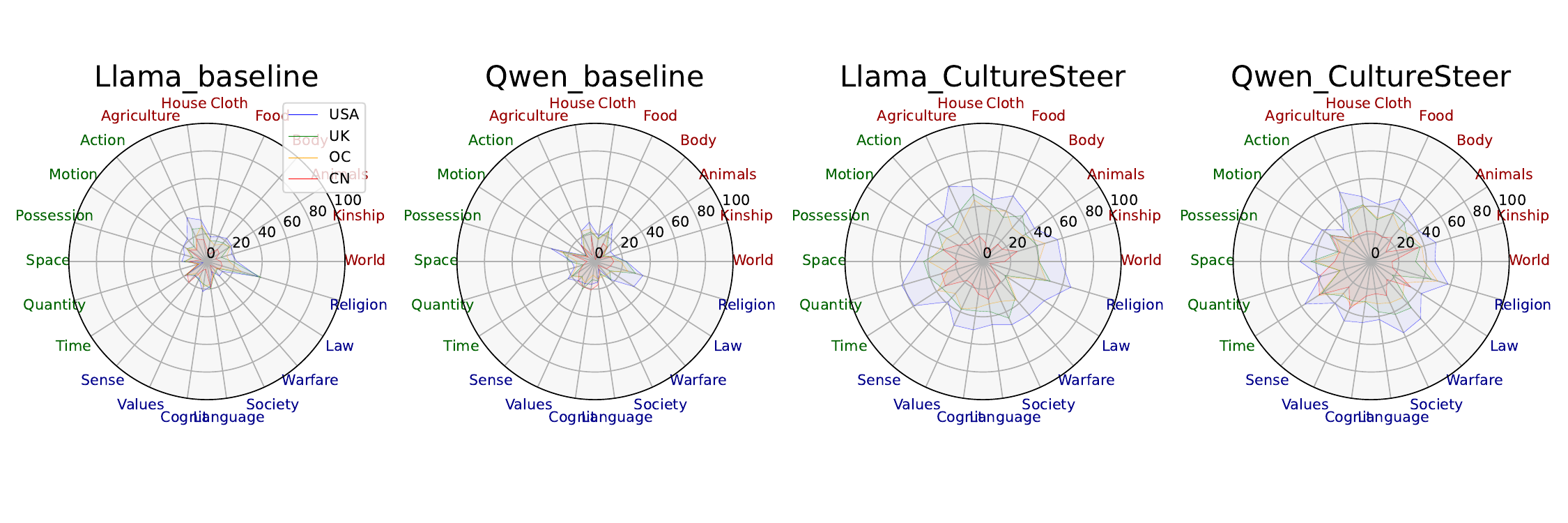}
    }
    \subfigure[Performance in PWR@10]{
        \includegraphics[width=1\textwidth]{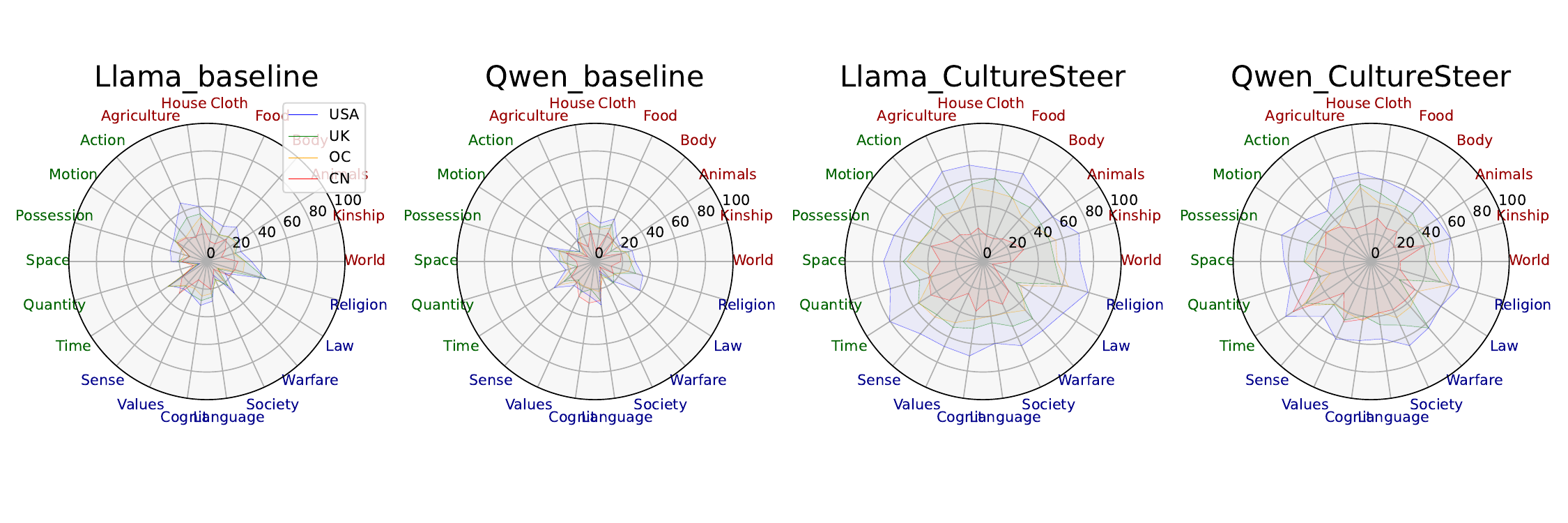}
    }
    \caption{Graunalr performance in PWR@3, 5 and 10.}
    \label{fig:radars}
\end{figure*}

The granular performance is shown in Figure~\ref{fig:radars}.
When K=3, 5, 10, the CultureSteer results consistently exhibit comparably weaker performance in Global Knowledge categories, while cultural disparities in Llama remain greater than those in Qwen.







\end{document}